\documentclass[10pt,twocolumn,letterpaper]{article}

\usepackage{iccv}
\usepackage{times}
\usepackage{epsfig}
\usepackage{graphicx}
\usepackage{amsmath}
\usepackage{amssymb}

\usepackage{subfigure}
\usepackage{pbox}
\usepackage{wrapfig}
\usepackage{chngcntr}

\usepackage{booktabs}
\usepackage{array}
\usepackage{xspace}

\usepackage[pagebackref=true,breaklinks=true,letterpaper=true,colorlinks,bookmarks=false]{hyperref}

\iccvfinalcopy

\newcommand{\mycaption}[2]{\caption{\textbf{#1.}\xspace#2}}
\newcommand{\warpseg}[0]{NetWarp}
\newcommand{\bz}{\mathbf{z}}

\ificcvfinal\fi
\begin{document}

\title{Semantic Video CNNs through Representation Warping}

\author{
Raghudeep Gadde$^{1, 3}$, Varun Jampani$^{1, 4}$ and Peter V. Gehler$^{1, 2, 3}$\\
$^1$MPI for Intelligent Systems,~~~~~~~~ $^2$University of W{\"u}rzburg\\
$^3$Bernstein Center for Computational Neuroscience,~~~~~~~~ $^4$NVIDIA\\
\texttt{\{raghudeep.gadde,varun.jampani,peter.gehler\}@tuebingen.mpg.de}\\
}

\maketitle

\begin{abstract}
  \vspace{-0.4cm}
  In this work, we propose a technique to convert CNN models for semantic
  segmentation of static images into CNNs for video data. We describe a
  warping method that can be used to augment existing architectures with very
  little extra computational cost. This module is called \warpseg~and we demonstrate
  its use for a range of network architectures. The main design principle is to use
  optical flow of adjacent frames for warping internal network representations
  across time. A key insight of this work is that fast optical flow methods
  can be combined with many different CNN architectures for improved performance and
  end-to-end training.
  Experiments validate that the proposed approach incurs only little extra computational cost,
  while improving performance, when video streams are available.
  We achieve new state-of-the-art results on the CamVid and Cityscapes benchmark
  datasets and show consistent improvements over different baseline networks.
  Our code and models are available at \url{http://segmentation.is.tue.mpg.de}
\end{abstract}

\vspace{-0.5cm}
\section{Introduction}

It is fair to say that the empirical performance of semantic image segmentation
techniques has seen dramatic improvement in the recent years with the onset of
Convolutional Neural Network (CNN) methods. The driver of this development have been
large image segmentation datasets and the natural
next challenge is to develop fast and accurate video segmentation methods.

The number of proposed CNN models for semantic image segmentation by far outnumbers those for video data.
A naive way to use a single image CNN for video is to apply it frame-by-frame, effectively ignoring the temporal information altogether.
However, frame-by-frame application often yields to jittering across frames, especially at object boundaries.
Alternative approaches include the use of conditional random field (CRF) models on video data
to fuse the predicted label information across frames or the development of
tailored CNN architectures for videos. A separate CRF applied to the CNN
predictions has the limitation, that it has no access to internal representations of
the CNNs. Thus the CRF operates on a representations (the labels) that has already been condensed.
Furthermore, existing CRFs for video data are often too slow for practical purposes.

\begin{figure*}[t]
\begin{center}
\centerline{\includegraphics[width=2\columnwidth]{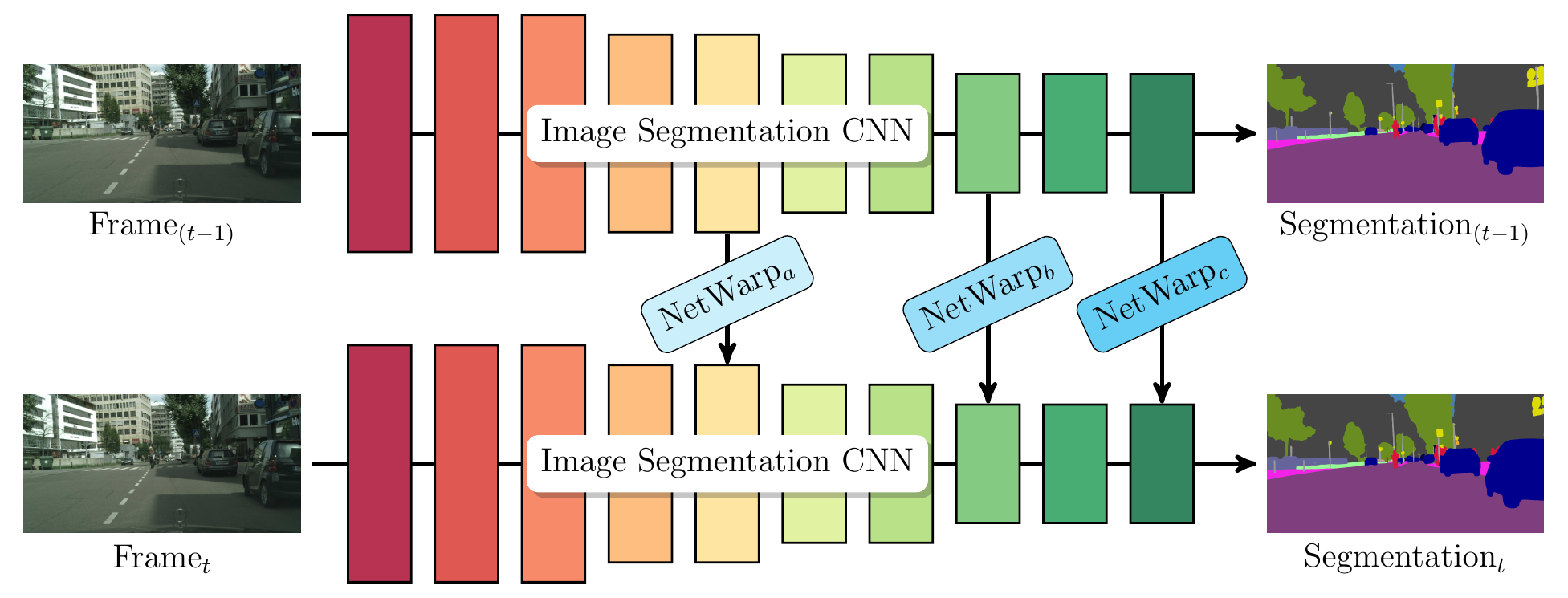}}
  \mycaption{Schematic of the proposed video CNN with~\warpseg~modules}{This illustration depicts the use of
  \warpseg~modules in three different layers of a image CNN. The video CNN
  is applied in an online fashion, looking back only one frame. The CNN filter activations for the current frame are modified   by the corresponding representations of the previous frame via \warpseg~modules.}
  \label{fig:teaser}
\end{center}
\vspace{-0.6cm}
\end{figure*}

We aim to develop a video segmentation technique that makes use of
temporal coherence in video frames and re-use strong single image segmentation CNNs.
For this, we propose a conceptually simple approach
to convert existing image CNNs into video CNNs that uses only very little
extra computational resources. We achieve this by `\warpseg', a
neural network module that warps the
intermediate CNN representations of the previous frame to the corresponding
representations of the current frame. Specifically, the \warpseg~module
uses the optical flow between two adjacent frames and then learns to
transform the intermediate CNN representations through an extra set of operations.
Multiple \warpseg~modules can be used at different layers of the CNN hierarchies
to warp deep intermediate representations across time, as depicted in Fig.~\ref{fig:teaser}.

Our implementation of \warpseg~takes only about 2.5 milliseconds to process an intermediate
CNN representation of $128 \times 128$ with $1024$ feature channels. It is
fully differentiable and can be learned using standard back propagation techniques
during training of the entire CNN network. In addition, the resulting video CNN model
with \warpseg~modules processes the frames in an online fashion, \ie, the system has access only to
the present and previous frames when predicting the segmentation of the present frame.

We augmented several existing state-of-the-art image segmentation CNNs
using \warpseg. On the current standard
video segmentation benchmarks of CamVid~\cite{brostow2009semantic} and
Cityscapes~\cite{Cordts2016Cityscapes}, we consistently observe
performance improvements in comparison to base network that is applied in
a frame-by-frame mode. Our video CNNs also outperformed other recently proposed
(CRF-)architectures and video propagation techniques setting up
a new state-of-the-art on both CamVid and Cityscapes datasets.

In Section~\ref{sec:related}, we discuss the related works on video segmentation.
In Section~\ref{sec:method}, we describe the \warpseg~module and how it is used to
convert image CNNs into video CNNs. In Section~\ref{sec:exps}, experiments
on CamVid and Cityscapes are presented. We conclude with a discussion
in Section~\ref{sec:conclusion}.

\section{Related Works}
\label{sec:related}

We limit our discussion of the literature on semantic segmentation
to those works concerning the video data.
Most semantic video segmentation approaches implement the strategy to
first obtain a single frame predictions using a classifier such as random forest
or CNN, and then propagate this information using CRFs or filtering techniques
to make the result temporally more consistent.

One possibility to address semantic video segmentation is by means of the 3D scene structure.
Some works~\cite{brostow2008segmentation,floros2012joint,sturgess2009combining}
build models that use 3D point clouds that have been obtained with structure from motion.
Based on these geometrical and/or motion features, semantic segmentation is improved.
More recent works~\cite{kundu2014joint,sengupta2013urban}
propose the joint estimation of 2D semantics and 3D reconstruction of the scenes
from the video data. While 3D information is very informative, it is also costly to
obtain and comes with prediction errors that are hard to recover from.

A more popular route~\cite{ess2009segmentation,chen2011temporally,de2012line,miksik2013efficient,tripathi2015semantic,
kundu2016feature,liu2015multi} is to construct large graphical models that connect different
video pixels to achieve temporal consistency across frames.
The work of~\cite{de2012line} proposes a Perturb-and-MAP random field model with
spatio-temporal energy terms based on Potts model.~\cite{chen2011temporally} used dynamic
temporal links between the frames but
optimizes for a 2D CRF with temporal energy terms. A 3D dense CRF across video
frames is constructed in~\cite{tripathi2015semantic} and optimized using mean-field approximate
inference. The work of~\cite{liu2015multi} proposed a joint model for predicting semantic labels for
supervoxels, object tracking and geometric relationship between the objects.
Recently,~\cite{kundu2016feature} proposed a technique for optimizing the feature spaces
for 3D dense CRF across video pixels. The resulting CRF model is applied on top of
the unary predictions obtained with CNN or some other techniques.
In~\cite{hur2016joint}, a joint model to estimate both optical flow
and semantic segmentation is designed.~\cite{lei2016recurrent} proposed a CRF model and an effiecient
inference technique to fuse CNN unaries with long range spatio-temporal cues estimated by
recurrent temporal restricted Boltzmann machine.
We avoid the CRF construction and filter
the intermediate CNN representations directly. This results in fast runtime and a natural
way to train any augmented model by means of gradient descent.

More related to our technique are fast filtering techniques. For example, ~\cite{miksik2013efficient}
learns a similarity function between pixels of consecutive frames to propagate
predictions across time. The approach of~\cite{jampani2017video} implements a neural network
that uses learnable bilateral filters~\cite{jampani_2016_cvpr}
for long-range propagation of information across video frames.
These filtering techniques propagate information after the semantic labels are computed
for each frame, whereas in contrast, our approach does filtering based propagation across
intermediate CNN representations making it more integrated into CNN training.

\begin{figure*}[t]
\begin{center}
\centerline{\includegraphics[width=2\columnwidth]{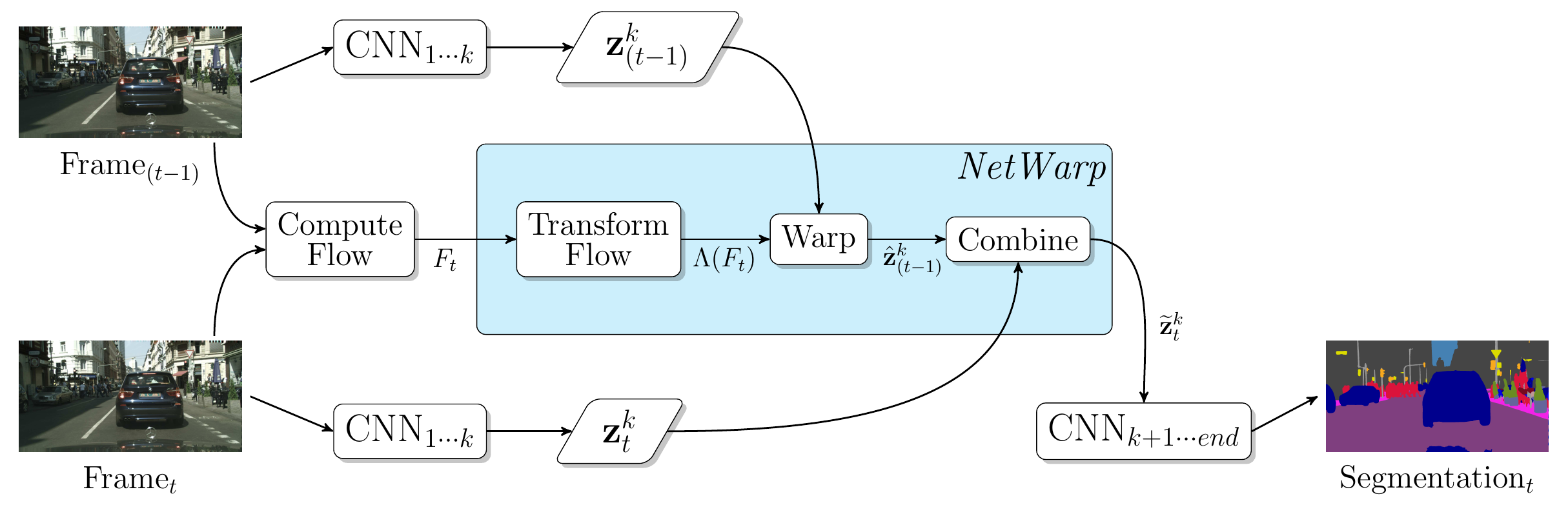}}
  \mycaption{Illustration of computations in a~\warpseg~module}{First, optical flow $F_t$
  is computed between two video frames at time steps $t$ and $t-1$.
  Then the \warpseg~module transforms the flow $\Lambda(F_t)$ with few convolutional layers;
  warps the activations $\bz^k_{(t-1)}$ of the previous frame and and combines the
  warped representations with those of the present frame $\bz^k_t$. The resulting representation
  $\widetilde{\bz}^k_t$ is then passed onto the remaining CNN layers for semantic segmentation.
  }
  \label{fig:warpseg}
\end{center}
\vspace{-1.0cm}
\end{figure*}

The use of CNNs (\eg,~\cite{long2014fully,chen2014semantic}) resulted in a surge of
performance in semantic segmentation. But, most CNN techniques work on single images.
The authors of~\cite{shelhamer2016clockwork} observed that the semantics change slowly across time
and re-use some intermediate representations from
the previous frames while computing segmentation for the present frame. This results
in faster runtime but a loss in accuracy. In contrast, our approach uses
adjacent frame deep representations for consistent predictions across frames
resulting in improved prediction accuracy.

Although several works proposed neural network approaches for processing several
video frames together, they are mostly confined to video level tasks such as
classification or captioning. The works of~\cite{ji20133d,karpathy2014large}
use 3D convolutions across frames for action recognition.
In~\cite{donahue2015long}, LSTMs are used in a recurrent
network for recognition and captioning. Two stream optical flow and image
CNNs~\cite{simonyan2014two,wang2016temporal,liu2015multi} are among the
state-of-the-art approaches for visual action recognition. Unlike video level tasks,
pixel-level semantic video segmentation requires filtering at pixel-level.
This work proposes a way of doing local information propagation across video frames.

A related task to semantic video segmentation is video object segmentation.
Like in semantic video segmentation literature, several works~\cite{reso2014interactive,
li2005video,price2009livecut,wang2005interactive,jain2014supervoxel} aim to
reduce the complexity of graphical model structure with spatio-temporal superpixels.
Some other works use nearest neighbor fields~\cite{fan2015jumpcut} or optical
flow~\cite{chuang2002video} for estimating correspondence between different
frame pixels. These works use pixel correspondences across frames to refine or
propagate labels, whereas the proposed approach refines the intermediate
CNN representations with a module that is easy to integrate into current CNN
frameworks.

\section{Warping Image CNNs to Video CNNs}
\label{sec:method}

Our aim is to convert a given CNN network architecture, designed to work on single images
into a segmentation CNN for video data.
Formally, given a sequence of $n$ video frames denoted as
$I_1,I_2,\cdots,I_n$, the task is to predict semantic segmentation for every video frame.
Our aim is to process the video frames
online, \ie, the system has access only to previous frames when predicting
the segmentation of the present frame.

The main building block is a~\warpseg~module that warps the
intermediate ($k^{th}$ layer) CNN representations $\mathbf{z}^k_{t-1}$ of the previous frame
and then combines with those of the present frame $\mathbf{z}^k_{t}$, where
$\mathbf{z}^1_t, \mathbf{z}^2_t, \cdots, \mathbf{z}^m_t$ denote the intermediate representations
of a given image CNN with $m$ layers.

\vspace{-0.4cm}
\paragraph{Motivation}
The design of the \warpseg~module is based on two specific insights from the recent
semantic segmentation literature. The authors of~\cite{shelhamer2016clockwork} showed
that intermediate CNN representations change only slowly over adjacent frames, especially
for deeper CNN layers. This inspired
the design of the clockwork convnet architecture~\cite{shelhamer2016clockwork}.
In~\cite{gadde16bilateralinception}, a bilateral inception module is constructed to
average intermediate CNN representations for locations across the image that are spatially
and photometrically close. There, the authors use super-pixels based on runtime
considerations and demonstrated improved segmentation results when applied to different
CNN architectures. Given these findings, in this work, we ask the question:
\emph{Does the combination of temporally close representations also leads to
more stable and consistent semantic predictions?}

We find a positive answer to this question. Using pixel correspondences, provided
by optical flow, to combine intermediate CNN representations of adjacent frames
consistently improves semantic predictions for a number of CNN architectures.
Especially at object boundaries and thin object structures, we observe a solid improvement.
Further, this warping can be performed at different layers in CNN architectures, as illustrated in
Fig.~\ref{fig:teaser} and incurs only a tiny extra computation cost to the entire
pipeline.

\subsection{\warpseg}

The \warpseg~module consists of multiple separate steps,
a flowchart overview is depicted in Fig.~\ref{fig:warpseg}.
It takes as input, an estimate of dense optical flow field and then performs
1. flow transformation, 2. representation warping, and 3. combination of
representations. In the following, we will first discuss the optical
flow computation followed by the description of each of the three separate steps.

\vspace{-0.4cm}
\paragraph{Flow Computation}
We use existing optical flow algorithms to obtain dense pixel
correspondences (denoted as $F_t$) across frames.
We chose a particular fast optical flow method to keep the runtime small.
We found that DIS-Flow~\cite{kroeger2016disflow}, which takes only about 5ms to compute
flow per image pair (with size 640 $\times$ 480) on a CPU, works well for
our purpose. Additionally, we experimented with the more accurate but slower
FlowFields~\cite{bailer15flowfields} method that requires several seconds per image pair to compute flow.
Formally, given an image pair, $I_t$ and $I_{(t-1)}$, the optical flow algorithm computes
the pixel displacement $(u,v)$ for every pixel location $(x,y)$ in $I_t$ to the
spatial locations $(x',y')$ in $I_{(t-1)}$. That is, $(x',y') = (x+u,y+v)$.
$u$ and $v$ are floating point numbers and represent pixel displacements
in horizontal and vertical directions respectively. Note that we compute the
\emph{reverse flow} mapping the present frame locations to locations in
previous frame as we want to warp previous frame representations.

\vspace{-0.4cm}
\paragraph{Flow Transformation}
Correspondences obtained with traditional optical flow methods might not
be optimal for propagating representations across video frames.
So, we use a small CNN to transform the pre-computed optical flow, which we
will refer to as FlowCNN and denote the transformation as $\Lambda(F_t)$.
We concatenate the original
two channel flow, the previous and present frame images, and the difference of the two frames.
This results in a 11 channel tensor as an input to the FlowCNN. The network itself
is composed of 4 convolution layers interleaved with ReLU nonlinearities.
All the convolution layers are made up of 3 $\times$ 3 filters and the number of
output channels for the first 3 layers are 16, 32 and 2 respectively.
The output of the third layer is then concatenated (skip connection) with the
original pre-computed flow which is then passed on to the last 3 $\times$ 3
convolution layer to obtain final transformed flow. This network architecture
is loosely inspired from the residual blocks in ResNet~\cite{he2016deep} architectures.
Other network architectures are conceivable. All parameters of FlowCNN are learned
via standard back-propagation. Learning is done on semantic segmentation only and
we do not include any supervised flow data as we are mainly interested in
semantic video segmentation. Figure~\ref{fig:transformed_flow} in the experimental section
shows how the flow transforms with the FlowCNN. We observe significant transformations
in the original flow with the FlowCNN and we will discuss more about these
changes in the experimental section.

\vspace{-0.4cm}
\paragraph{Warping Representations}

The FlowCNN transforms a dense correspondence field from
frame $I_t$ to the previous frame $I_{(t-1)}$.
Let us assume that we want to apply the \warpseg~module on the $k^{th}$ layer of the image CNN and
the filter activations for the adjacent frames are $\mathbf{z}^k_t$
and $\mathbf{z}^k_{(t-1)}$ (as in Fig~\ref{fig:warpseg}).
For notational convenience, we drop $k$ and refer to these
as $\bz_t$ and $\bz_{(t-1)}$ respectively.
The representations of the previous frame $\mathbf{z}_{(t-1)}$ are warped
to align with the corresponding present frame representations:
\begin{equation}
\hat{\mathbf{z}}_{(t-1)} = Warp(\mathbf{z}_{(t-1)},\Lambda(F_t)),
\end{equation}
where $\hat{\mathbf{z}}_{(t-1)}$ denotes the warped representations,
$F_t$ is the dense correspondence field and $\Lambda(\cdot)$ represents
the FlowCNN described above. Lets say we want to compute the warped representations
$\hat{\mathbf{z}}_{(t-1)}$ at a present frame's pixel location $(x,y)$ which is mapped to the
location $(x',y')$ in the previous frame by the transformed flow.
We implement $Warp$ as a bilinear interpolation of $\mathbf{z}_{(t-1)}$ at the
desired points $(x',y')$. Let $(x_1,y_1)$, $(x_1,y_2)$, $(x_2,y_1)$
and $(x_2,y_2)$ be the corner points of the previous frame's grid cell
where $(x',y')$ falls. Then the warping of $\mathbf{z}_{(t-1)}$ to obtain
$\hat{\mathbf{z}}_{(t-1)}(x,y)$ is given as standard bilinear interpolation:

\begin{align}
\begin{split}
& \hat{\mathbf{z}}_{(t-1)}(x,y) = \mathbf{z}_{(t-1)}(x',y') \\
&= \frac{1}{\eta} \begin{bmatrix} x_2-x' \\ x'-x_1 \end{bmatrix}^\top \begin{bmatrix} \mathbf{z}_{(t-1)}(x_1,y_1) & \mathbf{z}_{(t-1)}(x_1,y_2) \\ \mathbf{z}_{(t-1)}(x_2,y_1)& \mathbf{z}_{(t-1)}(x_2,y_2) \end{bmatrix} \begin{bmatrix} y_2-y' \\ y'-y_1 \end{bmatrix}
\end{split}
\end{align}
where $\eta = 1/(x_2-x_1)(y_2-y_1)$.
In case $(x',y')$ lies outside the spatial domain of $\mathbf{z}_{(t-1)}$, we
back-project $(x',y')$ to the nearest border in $\mathbf{z}_{(t-1)}$. The above
warping function is differentiable at all the points except when the flow values
are integer numbers. Intuitively, this is because the the corner points used for the
interpolation suddenly change when $(x',y')$ moves across from one grid cell to
another. To avoid the non-differentiable case, we add a small $\epsilon$ of 0.0001
to the transformed flow. This makes the warping module differentiable
with respect to both the previous frame representations and the transformed flow.
We implement gradients using standard matrix calculus.
 Due to strided pooling, deeper CNN representations are typically of smaller resolution in comparison to the image
 signal. The same strides are used for the transformed optical flow to obtain the pixel
 correspondences at the desired resolution.

 \begin{table*}[th!]
     \scriptsize
     \centering
     \begin{tabular}{p{3.4cm}>{\centering\arraybackslash}p{1.0cm}>{\centering\arraybackslash}
       p{1.0cm}>{\centering\arraybackslash}p{1.0cm}>{\centering\arraybackslash}p{1.0cm}>{\centering\arraybackslash}p{1.0cm}
       >{\centering\arraybackslash}p{1.0cm} >{\centering\arraybackslash}p{1.0cm}>{\centering\arraybackslash}p{1.0cm}
       >{\centering\arraybackslash}p{1.3cm}}
         \toprule
         \scriptsize
         PlayData-CNN (IoU: 68.9) & Conv1\_2 & Conv2\_2 & Conv3\_3 & Conv4\_3 & Conv5\_3 & FC6 & FC7 & FC8 & FC6 + FC7\\ [0.1cm]
         \midrule
         +~\warpseg~(without FlowCNN) & 69.3 & 69.5 & 69.5 & 69.4 & 69.5 & 69.6 & 69.4 & 69.3 & \textbf{69.8}\\
         +~\warpseg~(with FlowCNN) & 69.6 & 69.6 & 69.6 & 69.5 & 69.7 & 69.8 & 69.7 & 69.5 & \textbf{70.2}\\
         \bottomrule
         \\
     \end{tabular}
     \mycaption{The effect of where~\warpseg~modules are inserted}{Shown are test IoU scores on CamVid for
     augmented versions of the PlayData-CNN. We observe an improvement (frame-by-frame results in 68.9 IoU)
     independent of where a \warpseg~is included. Refining the flow estimate typically leads to slightly better results.
     }
     \label{tbl:playdata-camvid-ablation}
     \vspace{-0.4cm}
 \end{table*}

\vspace{-0.4cm}
\paragraph{Combination of Representations}
Once the warped activations of the previous frame $\hat{\mathbf{z}}^k_{(t-1)}$
are computed with the above mentioned procedure, they are linearly combined
with the present frame representations $\mathbf{z}^k_{t}$
\begin{equation}
\widetilde{\mathbf{z}}^k_{t} = \mathbf{w}_1 \odot \mathbf{z}^k_{t} + \mathbf{w}_2 \odot \hat{\mathbf{z}}^k_{(t-1)},
\end{equation}
where $\mathbf{w}_1$ and $\mathbf{w}_2$ are weight vectors with the same length as
the number of channels in $\mathbf{z}^k$; and $\odot$ represents per-channel scalar
multiplication. In other words, each channel of the
frame $t$ and the corresponding channel of the warped representations in
the previous frame $t-1$ are linearly combined.
The parameters $\mathbf{w}_1, \mathbf{w}_2$ are learned via standard back-propagation.
The result $\widetilde{\mathbf{z}}^k_{t}$ is then passed on to the remaining image CNN layers.
Different computations in the~\warpseg~module are illustrated in Fig.~\ref{fig:warpseg}.

\vspace{-0.4cm}
\paragraph{Usage and Training}
The inclusion of \warpseg~modules still allows end-to-end
training. It can be easily integrated in different
deep learning architectures. Note that back-propagating a loss from frame $t$ will
affect image CNN layers (those preceding \warpseg~modules) for the present and also previous frames.
We use shared filter weights for the image CNN across the frames.
Training is possible also when ground truth label
information is available for only some and not all frames, which is generally the case.

Due to GPU memory constraints, we make an approximation and only use two frames at a time.
Filter activations from frame $t-1$ would receive updates from $t-2$ when unrolling the
architecture in time, but we ignore this effect because of the hardware memory limitations.
The \warpseg~module can be included at different depths and multiple \warpseg~modules
can be used to form a video CNN. In our experiments, we used the same flow transformation $\Lambda(\cdot)$
when multiple \warpseg~modules are used.
We used the Adam~\cite{adam} stochastic gradient descent method for optimizing
the network parameters. Combination weights are initialized with $\mathbf{w}_1 = 1$
and $\mathbf{w}_2 = 0$, so the initial video network is identical to the
single image CNN. Other \warpseg~parameters are initialized randomly with
a Gaussian noise.
All our experiments and runtime analysis were performed
using a Nvidia TitanX GPU and a 6 core Intel i7-5820K CPU clocked at 3.30GHz machine.
Our implementation that builds on the Caffe~\cite{jia2014caffe} framework is available at
\url{http://segmentation.is.tue.mpg.de}.

\section{Experiments}
\label{sec:exps}

We evaluated the ~\warpseg~modules using the two challenging semantic video segmentation
benchmarks for which video frames and/or annotations are available: CamVid~\cite{brostow2009semantic} and Cityscapes~\cite{Cordts2016Cityscapes}.
Both datasets contain real world video sequences of street scenes. We choose different
popular CNN architectures of ~\cite{yu2016multi,richter2016playing,zhao2016pyramid} and augmented
them with the \warpseg~modules at different places across the network.
We follow standard protocols and report the standard Intersection over Union (IoU) score which is defined in terms
of true-positives (TP), false-positives (FP) and false-negatives (FP): ``TP / (TP + FP + FN)" and additionally
the instance-level iIoU for Cityscapes~\cite{Cordts2016Cityscapes}.
We are particularly interested in the segmentation effects around the boundaries. Methods
like the influential DenseCRFs~\cite{krahenbuhl2011efficient} are particularly good
in segmentation of the object boundaries. Therefore, we adopt the methodologies
from~\cite{kohli2009robust,krahenbuhl2011efficient} and measure the IoU performance only in a
narrow band around the ground truth label changes (see Fig.17 in~\cite{kohli2009robust}).
We vary the width of this band and
refer to this measure as trimap-IoU (tIoU). In all the experiments, unless specified,
we use a default trimap band of 2 pixels.

\vspace{-0.2cm}
\subsection{CamVid Dataset}
The CamVid dataset contains 4 videos with ground-truth labelling available for every
30th frame. Overall, the dataset has 701 frames with ground-truth. For direct comparisons
with previous works, we used the same train, validation and test splits
as in~\cite{yu2016multi,kundu2016feature,richter2016playing}. In all our models, we use only
the \emph{train} split for training and report the performance on the \emph{test} split.
We introduced~\warpseg~modules in two popular segmentation CNNs for this dataset: One is
PlayData-CNN from~\cite{richter2016playing} and another is Dilation-CNN from~\cite{yu2016multi}.
Unless otherwise mentioned, we used DIS-Flow~\cite{kroeger2016disflow} for the experiments on this dataset.

With PlayData-CNN~\cite{richter2016playing} as the base network,
we first study how the~\warpseg~module performs when introduced at different stages of the network.
The network architecture of PlayData-CNN is made of five convolutional blocks, each with 2 or 3 convolutional
layers, followed by three 1$\times$1 convolution layers (FC layers). We add the~\warpseg~module
to the following layers at various depths of the network: \textit{Conv$_{1\_2}$},
\textit{Conv$_{2\_2}$}, \textit{Conv$_{3\_3}$}, \textit{Conv$_{4\_3}$}, \textit{Conv$_{5\_3}$},
\textit{FC$_6$}, \textit{FC$_7$} and \textit{FC$_8$} layers. The corresponding IoU scores
are reported in Tab.~\ref{tbl:playdata-camvid-ablation}. We find a consistent improvement over the PlayData-CNN
performance of 68.9\% irrespective of the \warpseg~locations in the network. Since~\warpseg~modules at \textit{FC$_6$} and \textit{FC$_7$} performed
slightly better, we chose to insert~\warpseg~at both the locations in our final model with this
base network. We also observe consistent increase in IoU with the use of flow transformation
across different~\warpseg~locations. Our best model with two~\warpseg~modules yields an IoU
score of 70.2\% which is a new state-of-the-art on this dataset.
Adding~\warpseg~modules to CNN introduces very few additional parameters.
For example, two \warpseg~modules
at \textit{FC$_6$} and \textit{FC$_7$} in the PlayData-CNN have about 16K parameters, a mere 0.012\% of all
134.3M parameters. The experiments in Tab.~\ref{tbl:playdata-camvid-ablation} indicate that improvements
can be contributed to the temporal information propagation at multiple-depths. As a baseline,
concatenating corresponding \textit{FC$_6$} and \textit{FC$_7$} features across
frames
resulted in only 0.1\% IoU improvement compared to 1.3\% using~\warpseg~modules.

In Tab.~\ref{tbl:playdatacnn-camvid-diffflows}, we show the effect of using
a more accurate but slower optical flow method of Flowfields~\cite{bailer15flowfields}.
Results indicate that there is only a 0.1\% improvement in IoU with Flowfields but this incurs
a much higher runtime for flow computation. Results also indicate that our approach
outperformed the current state-of-the-art approach of VPN from~\cite{jampani2017video}
by a significant margin of 0.8\% IoU, while being faster.

\begin{figure}[t]
\begin{center}
\centerline{\includegraphics[width=\columnwidth]{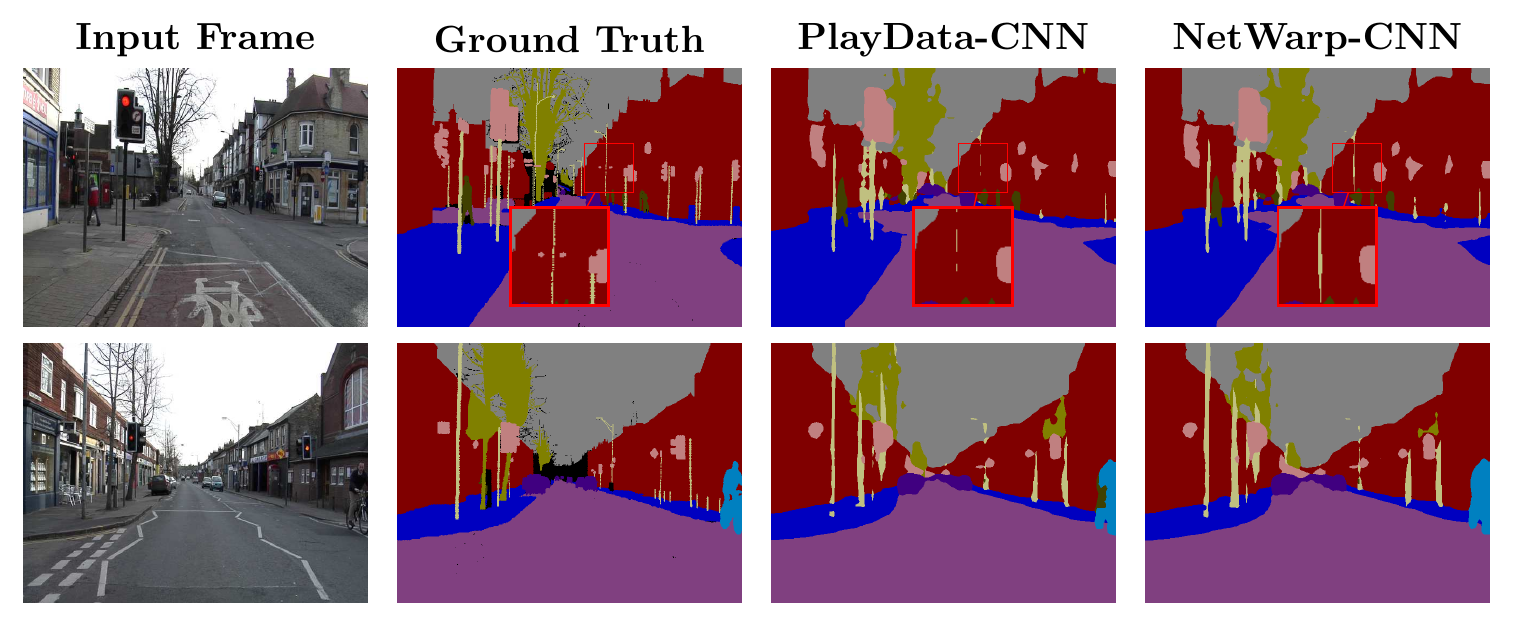}}
  \mycaption{Qualitative results from the CamVid dataset}{Notice how~\warpseg-CNN
  recovers some thin structures in the top row and corrects some regions (on cyclist) in the second row.}
  \label{fig:camvid_img_visuals}
  \vspace{-0.9cm}
\end{center}
\end{figure}

\begin{table}[t]
    \scriptsize
    \centering
    \begin{tabular}{p{3.3cm}>{\centering\arraybackslash}p{0.8cm}>{\centering\arraybackslash}
      p{0.8cm}} 
        \toprule
        \scriptsize
        & IoU & tIoU\\[0.1cm] 
        \midrule
        PlayData-CNN~\cite{richter2016playing} & 68.9 & 39.0  \\
        +~\warpseg~ (with DIS-Flow~\cite{kroeger2016disflow}) & 70.2 & 39.9 \\
        +~\warpseg~ (with FlowFields~\cite{bailer15flowfields}) & \textbf{70.3} & \textbf{40.1} \\
        \midrule
        + VPN~\cite{jampani2017video} & 69.5 & - \\ 
        \bottomrule
        \\
    \end{tabular}
    \mycaption{CamVid Results using PlayData-CNN}
    {Shown are the IoU and tIoU scores from different methods using a fast flow from DIS\cite{kroeger2016disflow} and an accurate flow from~\cite{bailer15flowfields} for~\warpseg~augmented PlayData-CNN.
    \label{tbl:playdatacnn-camvid-diffflows}}
    \vspace{-0.3cm}
\end{table}

\begin{table}[t]
    \scriptsize
    \centering
    \begin{tabular}{p{2.0cm}>{\centering\arraybackslash}p{0.8cm}>{\centering\arraybackslash}
      p{1.0cm}>{\centering\arraybackslash}p{1.5cm}}
        \toprule
        \scriptsize
        & IoU & tIoU & Runtime (ms)\\ [0.1cm]
        \midrule
        Dilation-CNN~\cite{yu2016multi} & 65.3 & 34.7 & 380\\
        +~\warpseg (Ours) & \textbf{67.1} & \textbf{36.6} & 395\\
        \midrule
        + FSO-CRF~\cite{kundu2016feature} & 66.1 & - & $>$ 10k\\
        + VPN~\cite{jampani2017video} & 66.7 & 36.1 & 680\\
        \bottomrule
        \\
    \end{tabular}
    \mycaption{CamVid Results using Dilation-CNN}
    {IoU, tIoU scores and runtimes (in milliseconds) for different methods.}
    \label{tbl:dilationcnn-camvid}
    \vspace{-0.3cm}
\end{table}

As a second network, we choose the Dilation-CNN from~\cite{yu2016multi}. This network consists of a standard CNN
followed by a context module with 8 dilated convolutional layers. For this network, we apply the \warpseg~module
on the output of each of these 8 dilated convolutions. Table~\ref{tbl:dilationcnn-camvid} shows the performance
and runtime comparisons  with the dilation CNN and other related techniques. With a runtime increase of 15 milliseconds,
we observe significant improvements in the order of 1.8\% IoU. The runtime increase assumes that the result of the
previous frame is already computed, which is the case for video segmentation.

\subsection{Cityscapes Dataset}

The Cityscapes dataset~\cite{Cordts2016Cityscapes}
comes with a total of 5000 video sequences of high-quality images (2048$\times$1024 resolution), partitioned
into 2975 train, 500 validation and 1525 test sequences.
The videos are captured in different weather conditions across 50 different cities in Germany
and Switzerland. In addition to the IoU and tIoU performance metrics, we report the
instance-level IoU score. Since IoU score is dominated
by large objects/regions (such as road) in the scene, the makers of this dataset proposed the iIoU
score that takes into account the relative size of different objects/regions. The iIoU score
is given as iTP/(iTP + FP + iFN), where iTP and iFN are the modified true-positive and
false-negative scores which are computed by weighting the contribution of each pixel by the ratio
of the average class instance size to the size of the respective ground truth instance. This measures
how well the semantic labelling represents the individual instances in the scene. For more details
on this metric, please refer to the original dataset paper~\cite{Cordts2016Cityscapes}.
For this dataset, we used DIS-Flow~\cite{kroeger2016disflow} for all networks augmented with \warpseg~modules.

We choose the recently proposed Pyramid Scene Parsing Network (PSPNet) from
\cite{zhao2016pyramid}. Because of high-resolution images in Cityscapes and GPU memory
limitations, PSPNet is applied in a sliding window fashion with a window size of 713$\times$713.
To achieve higher segmentation performance, the authors of~\cite{zhao2016pyramid} also evaluated a multi-scale version.
Applying the \emph{same} PSPNet on 6 different scales of an input image results to an improvement of 1.4\% IoU over the single-scale variant.
This increased performance comes at the cost of increased runtime. In the single-scale setting,
the network is evaluated on 8 different windows to get a full image result, whereas in the
multi-scale setting, the network is evaluated 81 times leading to 10 times increase in runtime.
We refer to the single-scale and multi-scale evaluations as PSPNet-SSc and PSPNet-MSc respectively.

The architecture of PSPNet is a ResNet101~\cite{he2016deep} network variant with pyramid style pooling layers.
We insert \warpseg~modules before and after the pyramid pooling layers.
More precisely, \warpseg~modules are added on both the \textit{Conv$_{4\_23}$} and \textit{Conv$_{5\_4}$}
representations. The network is then fine-tuned with 2975 train video sequences without any data augmentation.
We evaluate both the single-scale and multi-scale variants.
Table~\ref{tbl:pspnet-cityscapes} summarizes the quantitative results with and without \warpseg~modules, on validation
data scenes of Cityscapes.
We find an improvement of the IoU (by 1.2), tIoU (by 2.4) and iIoU (by 1.4) respectively over the single image PSPNet-SSc variant.
These improvements come with a low runtime overhead of 24 milliseconds.
Also the augmented multi-scale network improves all measures: IoU by 0.7, tIoU by 1.8, and iIoU by 1.4\%.

We chose to submit the results of the best performing method from the validation set
to the Cityscapes test server. Results are shown in Tab.~\ref{table:cityscapestestresults}. We find that
the \warpseg~augmented PSP multi-scale variant is on par with the current top performing method~\cite{wu2016wider} (80.5 vs. 80.6)
and out-performs current top models in terms of iIoU by a significant margin of 1.4\%.
In summary, at submission time, our result is best performing method in iIoU and
second on IoU\footnote{\url{https://www.cityscapes-dataset.com/benchmarks/}}.
Surprisingly, it is the only approach that uses video data. A closer look into class IoU score in
Tab.~\ref{table:cityscapestestresults} shows that our approach works particularly well for
parsing thin structures like poles, traffic lights, traffic signs etc. The improvement
is around 4\% IoU for the pole class.
Another observation is that adding~\warpseg~modules resulted in slight IoU performance decrease for
few broad object classes such as car, truck etc. However, we find consistent improvements across most of the
classes in other performance measures. The classwise iIoU scores that are computed on broad object classes like
car, bus etc show better performance on 7 out of 8 classes for~\warpseg. Further, analysing the classwise tIoU
scores on the validation set, we find that~\warpseg~performs better on 17 out of 19 classes.
Visual results in Fig.~\ref{fig:cityscapes_img_visuals} also indicate that the thin structures are better
parsed with the introduction of~\warpseg~modules.
Qualitatively, we find improved performance near boundaries compared to baseline CNN (see supplementary video\footnote{\url{https://youtu.be/T6gq5wMaxAo}}).
Visual results in Fig.~\ref{fig:cityscapes_img_visuals} also indicate that \warpseg~helps in rectifying the
segmentation of big regions as well.

\begin{table}[t]
    \scriptsize
    \centering
    \begin{tabular}{p{2.9cm}>{\centering\arraybackslash}p{0.6cm}>{\centering\arraybackslash}
      p{0.6cm}>{\centering\arraybackslash}p{0.6cm}>{\centering\arraybackslash}p{1.2cm}}
        \toprule
        \scriptsize
        & IoU & iIoU & tIoU & Runtime (s)\\ [0.1cm]
        \midrule
        PSPNet-SSc~\cite{zhao2016pyramid} & 79.4 & 60.7 & 39.7 & 3.00\\
        +\warpseg & \textbf{80.6} & \textbf{62.1} & \textbf{42.1} & 3.04\\
        \midrule
        PSPNet-MSc~\cite{zhao2016pyramid} & 80.8 & 62.2 & 41.5 & 30.3\\
        +\warpseg & \textbf{81.5} & \textbf{63.6} & \textbf{43.3} &30.5\\
        \bottomrule
        \\
    \end{tabular}
    \mycaption{Performance of PSPNet with \warpseg~modules on the Cityscapes validation dataset}
    {We observe consistent improvements with~\warpseg~modules across all three performance measures
    in both the single-scale (SSc) and multi-scale (MSc) settings, while only adding little time
    overhead.\label{tbl:pspnet-cityscapes}}
  \vspace{-0.5cm}
\end{table}

\begin{figure}[t]
\begin{center}
\centerline{\includegraphics[width=0.75\columnwidth]{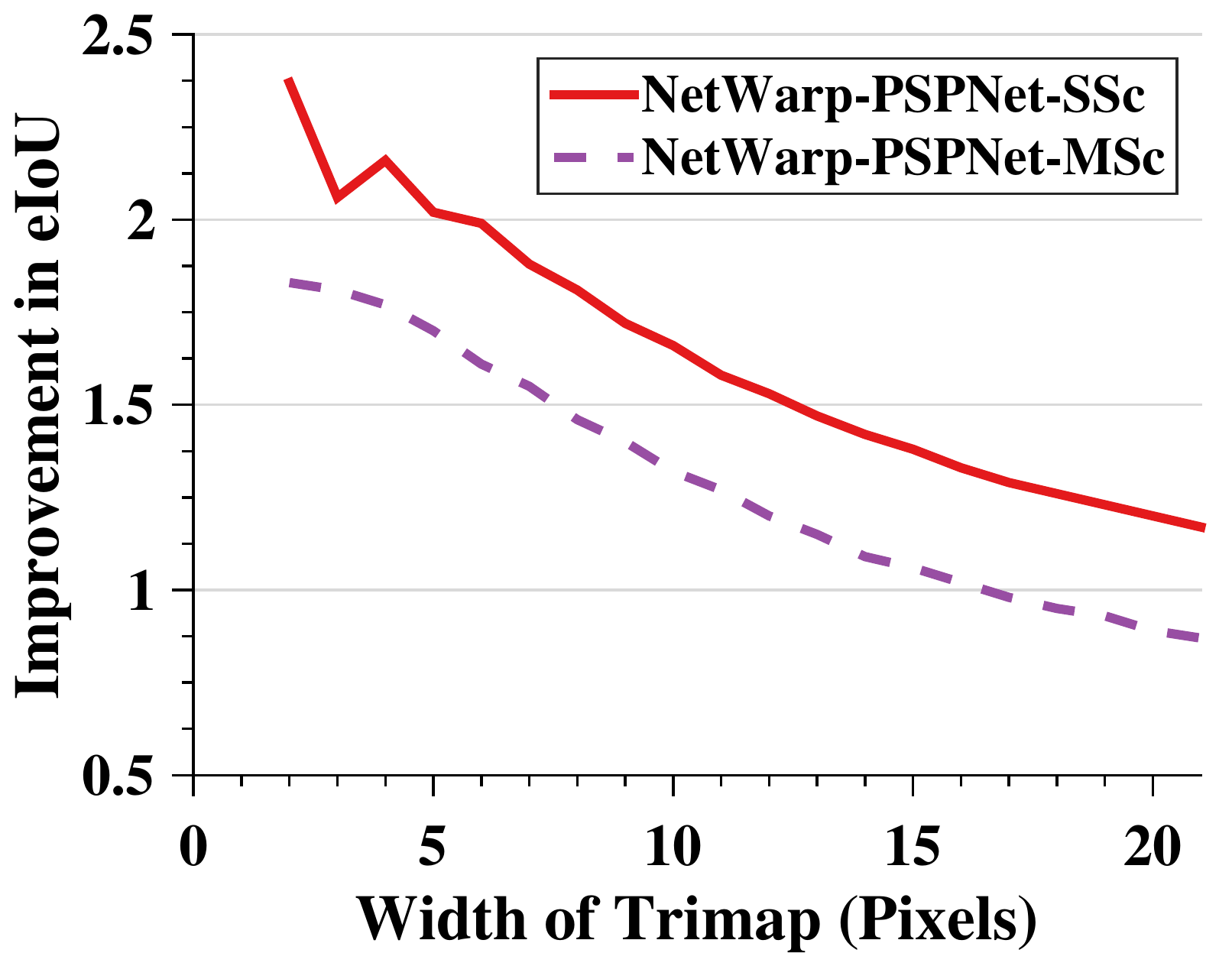}}
  \mycaption{tIoU improvement}{Relative improvements
  of IoU within trimaps as a function of trimap width. Most improvement is in regions close to object boundaries. }
  \label{fig:trimap_iou}
\end{center}
  \vspace{-1.1cm}
\end{figure}

In Fig.~\ref{fig:trimap_iou}, we show the relative improvement of the~\warpseg~augmented PSPNet
for different widths in the trimap-IoU. From this analysis, we can conclude that the IoU improvement is
especially due to better performance near object boundaries. This is true for both the
single and the multi-scale versions of the network.
Image CNNs, in general, are very good at segmenting large regions or objects like road
or cars. However, thin and fine structures are difficult to parse as information is
lost due to strided operations inside CNN. In part this is recovered by \warpseg~CNNs that
use the temporal context to recover thin structures. In Fig.~\ref{fig:cityscapes_img_visuals},
some qualitative results with static image CNN and our video CNN are shown. We observe that
the~\warpseg~module correct for thin structures but also frequently correct larger regions
of wrong segmentations. This is possible since representations for the same regions are
combined over different frames leading to a more robust classification.

\begin{figure}[t]
\begin{center}
\centerline{\includegraphics[width=\columnwidth]{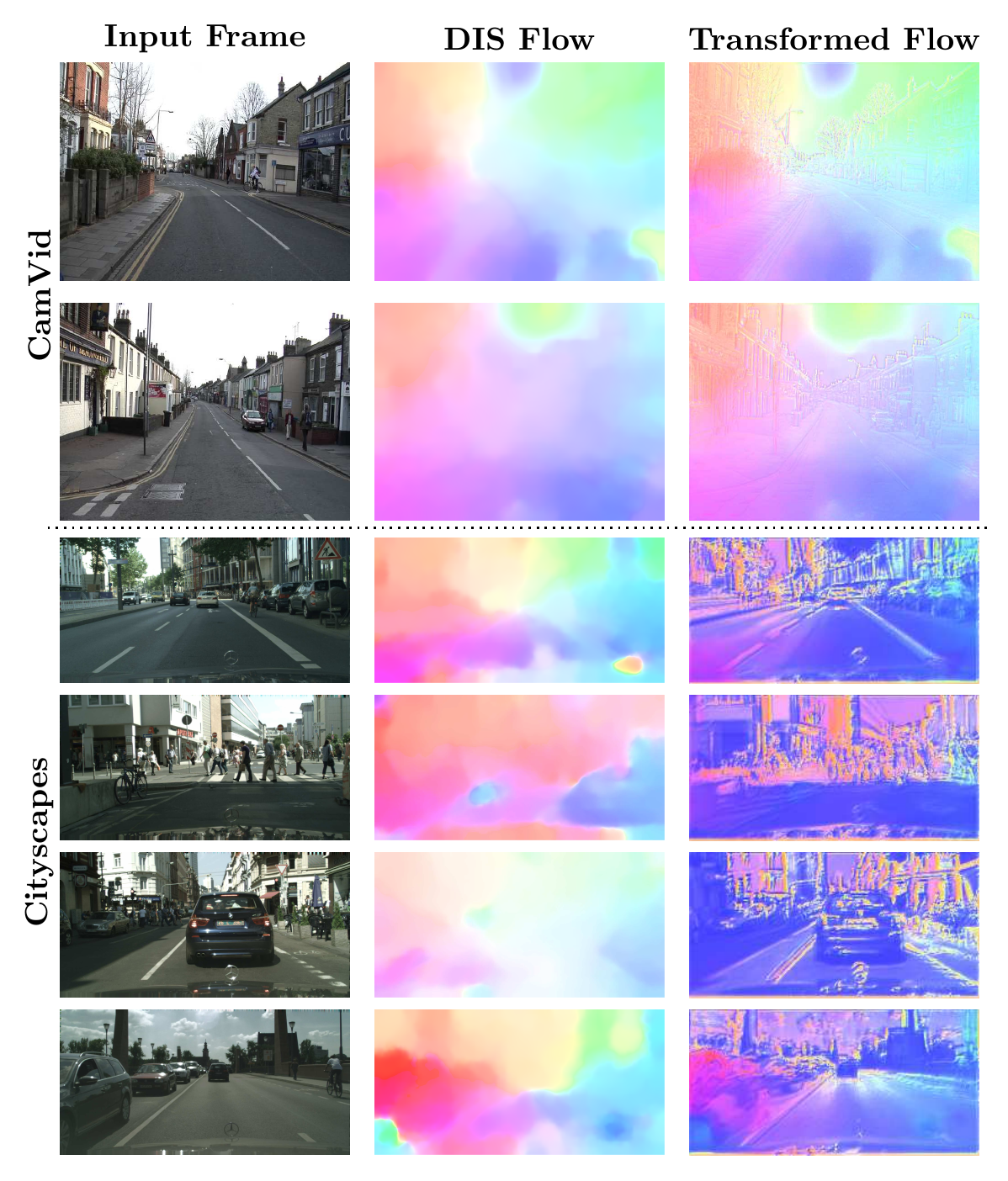}}
  \mycaption{Effect of flow transformation}{Example results of input and transformed flow, after training for semantic segmentation.
  There is a qualitative difference between CamVid and Cityscapes. Best
  viewed on screen.}
  \label{fig:transformed_flow}
\end{center}
  \vspace{-1.2cm}
\end{figure}

\begin{figure*}[t]
\begin{center}
\centerline{\includegraphics[width=2\columnwidth]{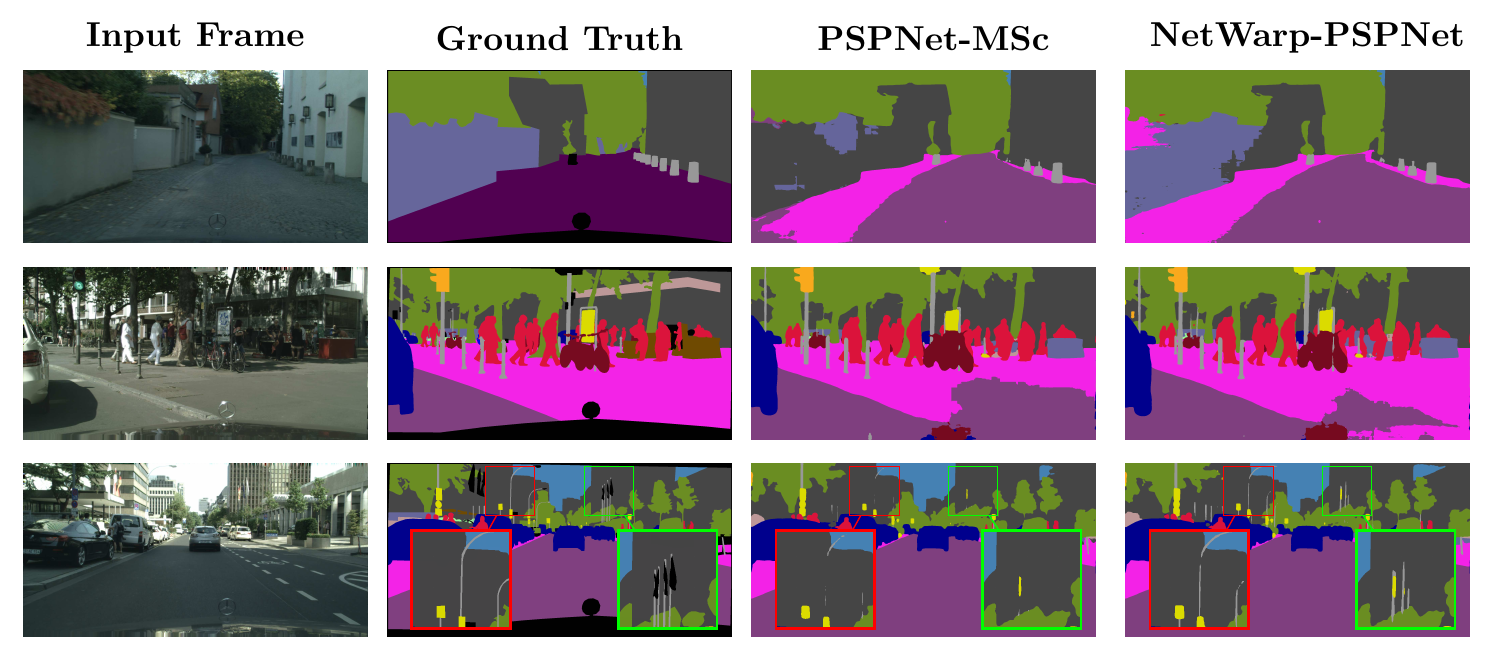}}
  \mycaption{Qualitative results from the Cityscapes dataset}{Observe how NetWarp-PSPNet
  is able to correct larger parts of wrong segmentation (top two rows) by warping
  activations across frames. The bottom row shows an example of improved segmentation of a
  thin structure. All results shown are obtained with the multi-scale versions.}
  \label{fig:cityscapes_img_visuals}
\end{center}
  \vspace{-0.7cm}
\end{figure*}

Next, we analyze how the DIS-Flow is transformed by the FlowCNN.
Figure~\ref{fig:transformed_flow} shows some visualizations of the transformed flow along
with the original DIS Flow fields. We can clearly observe that, in both CamVid and Cityscapes images,
the structure of the scene is much more pronounced in the transformed flow indicating that
the traditionally computed flow might not be optimal to find pixel correspondences
for semantic segmentation.

\begin{table*}[t]
  \scriptsize
  \centering
    \begin{tabular}{p{1.9cm}|p{0.5cm}>{\centering\arraybackslash}p{0.5cm}|>
{\centering\arraybackslash}p{0.25cm}>{\centering\arraybackslash}p{0.25cm}>
{\centering\arraybackslash}p{0.25cm}>{\centering\arraybackslash}p{0.25cm}>
{\centering\arraybackslash}p{0.25cm}>{\centering\arraybackslash}p{0.3cm}>
{\centering\arraybackslash}p{0.25cm}>{\centering\arraybackslash}p{0.25cm}>
{\centering\arraybackslash}p{0.25cm}>{\centering\arraybackslash}p{0.25cm}>
{\centering\arraybackslash}p{0.25cm}>{\centering\arraybackslash}p{0.25cm}>
{\centering\arraybackslash}p{0.25cm}>{\centering\arraybackslash}p{0.25cm}>
{\centering\arraybackslash}p{0.25cm}>{\centering\arraybackslash}p{0.25cm}>
{\centering\arraybackslash}p{0.25cm}>{\centering\arraybackslash}p{0.25cm}>
{\centering\arraybackslash}p{0.25cm}}
        \toprule
&\rotatebox[origin=c]{45}{iIoU}&\rotatebox[origin=c]{45}{IoU}&\rotatebox[origin=c]{45}{road}&\rotatebox[origin=c]{45}{sidewalk}&\rotatebox[origin=c]{45}{building}&\rotatebox[origin=c]{45}{wall}&\rotatebox[origin=c]{45}{fence}&\rotatebox[origin=c]{45}{pole}&\rotatebox[origin=c]{45}{trafficlight}&\rotatebox[origin=c]{45}{trafficsign}&\rotatebox[origin=c]{45}{vegetation}&\rotatebox[origin=c]{45}{terrain}&\rotatebox[origin=c]{45}{sky}&\rotatebox[origin=c]{45}{person}&\rotatebox[origin=c]{45}{rider}&\rotatebox[origin=c]{45}{car}&\rotatebox[origin=c]{45}{truck}&\rotatebox[origin=c]{45}{bus}&\rotatebox[origin=c]{45}{train}&\rotatebox[origin=c]{45}{motorcycle}&\rotatebox[origin=c]{45}{bicycle}\\[0.2cm]
      \midrule

PSPNet-MSc~\cite{zhao2016pyramid}&58.1&80.2&98.6&86.6&93.2&58.1&63.0&64.5&75.2&79.2&93.4&72.1&95.1&86.3&71.4&96.0&73.6&90.4&80.4&69.9&76.9\\
+\warpseg(Ours)&\textbf{59.5}&80.5&98.6&86.7&93.4&60.6&62.6&68.6&75.9&80.0&93.5&72.0&95.3&86.5&72.1&95.9&72.9&90.0&77.4&70.5&76.4\\
\midrule
ResNet-38~\cite{wu2016wider} &57.8&\textbf{80.6}&98.7&87.0&93.3&60.4&62.9&67.6&75.0&78.7&93.7&73.7&95.5&86.8&71.1&96.1&75.2&87.6&81.9&69.8&76.7\\
TuSimple~\cite{wang2017understanding}&56.9&80.1&98.6&85.9&93.2&57.7&61.2&67.2&73.7&78.0&93.4&72.3&95.4&85.9&70.5&95.9&76.1&90.6&83.7&67.4&75.7\\
LRR-4X~\cite{ghiasi2016laplacian}&47.9&71.9&98.0&81.5&91.4&50.5&52.7&59.4&66.8&72.7&92.5&70.1&95.0&81.3&60.1&94.3&51.2&67.7&54.6&55.6&69.6\\
RefineNet~\cite{lin2017RefineNet}&47.2&73.6&98.2&83.3&91.3&47.8&50.4&56.1&66.9&71.3&92.3&70.3&94.8&80.9&63.3&94.5&64.6&76.1&64.3&62.2&70.0\\
      \bottomrule
  \end{tabular}
  \mycaption{Results on the Cityscapes test dataset}{Average iIoU, IoU and class IoU scores of base PSPNet, with~\warpseg~modules and also other top performing methods taken from the Cityscapes benchmark website at the time of submission. The \warpseg~augmented PSPNet-MSc version achieves highest iIoU and is about on par with~\cite{wu2016wider} on IoU. Our video methods performs particularly well on small classes such as poles, traffic lights etc.\label{table:cityscapestestresults}}
\end{table*}

\vspace{-0.3cm}
\section{Conclusions and Outlook}
\label{sec:conclusion}

We presented \warpseg, an efficient and conceptually easy way to transform image CNNs
into video CNNs. The main concept is to transfer intermediate CNN filter activations of
adjacent frames based on transformed optical flow estimate.
The resulting video CNN is end-to-end trainable, runs in an online fashion and has only a small
computation overhead in comparison to the frame-by-frame application. Experiments on the current
standard benchmark datasets of Cityscapes and CamVid show improved performance for several
strong baseline methods. The final model sets a new state-of-the-art performance
on both Cityscapes and CamVid.

Extensive experimental analysis provide insights into the workings
of the~\warpseg~module. First, we demonstrate consistent performance improvements
across different image CNN hierarchies. Second, we find more temporally consistent
semantic predictions and better coverage of thin structures such as poles and traffic signs.
Third, we observed that the flow changed radically after
the transformation (FlowCNN) trained with the segmentation loss. From the qualitative results,
it seems that the optical flow at the object boundaries is important for semantic
segmentation. An interesting future work is to systematically study what properties of
optical flow estimation are necessary for semantic segmentation and the impact of different
types of interpolation in a~\warpseg~module.

Another future direction is to scale the video CNNs to use
multiple frames. Due to GPU memory limitations and to keep training easier, here,
we trained with only two adjacent frames at a time. In part this is due to the memory demanding
base models like ResNet101. Memory optimizing the CNN training would alleviate some
of the problems and enables training with many frames together. We also envision that the findings of this
paper are interesting for the design of video CNNs for different tasks other than
semantic segmentation.

{\noindent\small\textbf{Acknowledgements.} 
This work was supported by Max Planck ETH Center for Learning Systems and the research program of the Bernstein Center for
Computational Neuroscience, T{\"u}bingen, funded by the German Federal
Ministry of Education and Research (BMBF; FKZ: 01GQ1002). We thank Thomas Nestmeyer and Laura Sevilla for
their help with the supplementary.
}

{\small
\bibliographystyle{ieee}
\bibliography{references}
}

\end{document}